# Can AI detect pain and express pain empathy? A review from emotion recognition and a human-centered AI perspective

**Running head:** Pain Recognition and Empathic Expression in AI

Siqi Cao[1, 2, 3], Di Fu[1, 2, 5], Xu Yang[4], Stefan Wermter[5], Xun Liu[1, 2*], Haiyan Wu[6*]

[1]CAS Key Laboratory of Behavioral Science, Institute of Psychology, Chinese Academy of Sciences, Beijing, China

[2]Department of Psychology, University of Chinese Academy of Sciences, Beijing, China

[3]Department of Experimental Psychology, University of Oxford, Oxford, UK

[4]State Key Laboratory for Management and Control of Complex Systems, Institute of Automation, Chinese Academy of Sciences, Beijing, China

[5]Department of Informatics, University of Hamburg, Hamburg, Germany

[6]Centre for Cognitive and Brain Sciences and Department of Psychology, University of Macau, Taipa, Macau

## Declarations

### Funding

This work is partly supported by the National Natural Science Foundation of China (NSFC: U1736125, 62061136001, 31871142), the German Research Foundation (DFG) under project Transregio Crossmodal Learning (TRR-169), and SRG of the University of Macau.

### Conflicts of interest/Competing interests

All the authors declare no competing interests.

### Corresponding authors

Please address correspondence to Haiyan Wu (haiyanwu@um.edu.mo) or Xun Liu (liux@psych.ac.cn)

### Authors' contributions

H.W. conceived the presented topic. S.C. led the manuscript writing. X.Y., and S.W.. provided scientific suggestions for Section 2 and Section 3. D.F. and S.W. revised the writing structure. X.L. and H.W. supervised the writing process. All the authors contributed to the final manuscript.

## Abstract

Sensory and emotional experiences such as pain and empathy are essential for mental and physical health. Cognitive neuroscience has been working on revealing mechanisms underlying pain and empathy. Furthermore, as trending research areas, computational pain recognition and empathic artificial intelligence (AI) show progress and promise for healthcare or human-computer interaction. Although AI research has recently made it increasingly possible to create artificial systems with affective processing, most cognitive neuroscience and AI research do not jointly address the issues of empathy in AI and cognitive neuroscience. The main aim of this paper is to introduce key advances, cognitive challenges and technical barriers in computational pain recognition and the implementation of artificial empathy. Our discussion covers the following topics: How can AI recognize pain from unimodal and multimodal information? Is it crucial for AI to be empathic? What are the benefits and challenges of empathic AI? Despite some consensus on the importance of AI, including empathic recognition and responses, we also highlight future challenges for artificial empathy and possible paths from interdisciplinary perspectives. Furthermore, we discuss challenges for responsible evaluation of cognitive methods and computational techniques and show approaches to future work to contribute to affective assistants capable of empathy.

# 1 Introduction

Recent research has defined pain as an uncomfortable sensory and negative emotional experience with or without discernible tissue damage (Raja et al., 2020). From an evolutionary perspective, pain expression significantly affects interpersonal relationships (Williams, 2002). For instance, people show empathy for others in pain and are willing to help them relieve painful feelings (Lockwood et al., 2020). The challenges associated with pain recognition, in general, involve the following aspects: 1) A lack of cognitive, linguistic, and social abilities can hinder the accuracy of subjective assessments, particularly for young children and patients unable to express their feelings; 2) Self-reporting of pain has long been the predominant measure in medicine but if large-scale health monitoring is required, it can be unreliable and time-consuming. Most importantly, how reliable is the medical diagnosis of pain evaluation? It has been argued that clinical diagnosis is not rigorous and inefficient when subjectivity is involved. For example, depressed patients can display happiness briefly, but not in a genuinely happy mood, indicating that cues that are easy to overlook may mislead diagnosis.

Considering that clinical decisions are not necessarily reliable or objective to a certain extent, a new field in artificial pain recognition appears to emerge. A growing body of research examines the issue from the perspectives of cognitive neuroscience, philosophy, and human-centered artificial intelligence (AI). Over the past few decades, extensive research has been conducted in human behavioral sciences and neuroscience to determine the biological mechanisms of pain (Danziger et al., 2009; Krishnan et al., 2016; Lieberman & Eisenberger, 2009; Phelps et al., 2021; Smith et al., 2021). Research encompassing various disciplines has led to significant insights into pain recognition through a broad range of empirical and theoretical studies, laying the groundwork for future exploration of better interaction between humans and computers, humans and artificial intelligence, or humans and robots.

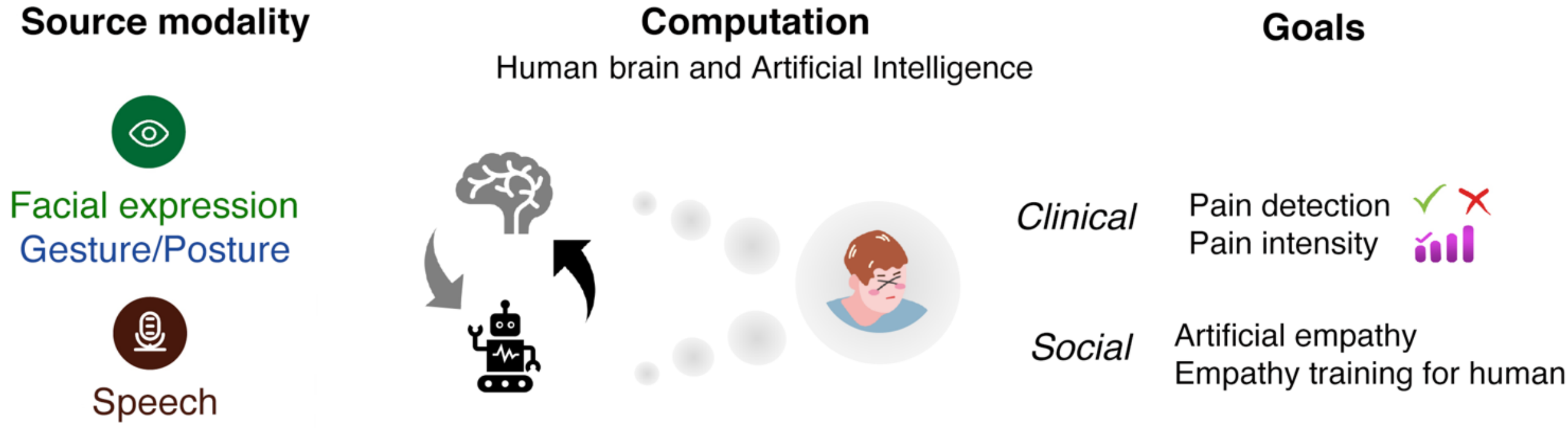

**Fig. 1** This review's overall illustration of topics, contents, and goals.

In this review, we intend to shed light on artificial empathy and training for empathy through computational pain recognition and to gain insight into what is needed to build a Human-In-The-Loop (HITL) system for pain recognition and artificial pain empathy. To begin with, we review three major modalities of pain recognition: face, voice, and body (Section 2; see Fig. 1 left). Then, in Section 3, cross-disciplinary ideas for existing and potential pain recognition practices are presented based on research in psychology and cognitive neuroscience (see Fig. 1 middle). While pain empathy plays a vital role in social functioning, computational pain recognition and artificial pain empathy have received little attention. Therefore, in Section 4, we present a growing body of knowledge concerning artificial pain empathy. We discuss the following questions in more detail: How can AI recognize pain from unimodal or multimodal inputs? Is it possible to contribute to this emerging field by studying human beings? Can we build an AI agent with empathy? Is empathic AI necessary? Our final discussion will also focus on ethical considerations in both research and prospective applications.

## 2 Unimodal pain recognition

Pain recognition can be divided into two questions: pain detection and pain intensity assessment. A pain detection problem is a binary classification that aims to determine whether a person is in pain. In light of considerations of precision, the focus of the research gradually shifted from pain detection to the estimation of pain intensity. Research on computational pain recognition has primarily focused on techniques based on unimodal input involving the face, voice, or body movement (Kaltwang et al., 2012; Oshrat et al., 2016; Semwal & Londhe, 2021). Furthermore, the temporal relationship has been explored, ranging from single static images to dynamic sequences of videos (Simon et al., 2008). As a starting point for pain recognition, we analyze several published pain databases based on a variety of modalities (see Supplementary Table 1). Each database collected pain expressions in different modes, mainly categorized into three types: acted, spontaneous, and elicited (acted expressions are from actors. Spontaneous expressions are from patients with chronic or acute pain. Elicited expressions indicate expressions evoked by audiovisual media or real interpersonal interaction). The collection of data in a laboratory setting is, however, controversial. Zhang et al. (2020) pointed out that people tend to change their biophysical signals to improve the extent to which their minds are being "read". Consequently, intentional noise leads to data instability and invalidity, making data collection challenging regardless of methods being used.

### 2.1 Face-based

Human faces are the most reliable and informative indicators of mental states, particularly emotions (Kaltwang et al., 2012) (see Supplementary Table 2). Researchers established the Facial Action Coding System (FACS) to identify different facial muscles (action units, AUs) that represent all kinds of emotions (Ekman & Friesen, 1978). Based on FACS, a well-established pain assessment standard with a 16-point scale was developed, known as Prkachin and Solomon Pain Intensity (PSPI) (Prkachin & Solomon, 2008). In addition, an Active Appearance Model (AAM) is also commonly used by researchers to identify facial patterns associated with spontaneous pain (Ashraf et al., 2007).

A growing concern has been raised regarding the ambiguity of pain, as some reports suggested that pain can sometimes appear as a combination of other emotions, such as disgust (Kunz et al., 2013). Identifying specific characteristics of pain is one of the most important aspects of pain recognition. Furthermore, whether methods frequently used for estimating pain intensity, such as regression, are stable has been controversial. For example, pain and eye closure have a positive correlation, while blinking and eye closure have similar facial characteristics when viewed independently. Pain intensity will be overestimated in images with closed eyes, increasing variance in continuous pain evaluations (Zhou et al., 2016). Future research needs to differentiate between video frame sequences according to their temporal relationship.

For many years, it has been considered time-consuming to code facial features in FACS-based video sequences. AU, PSPI, and AAMs provide the most comprehensive understanding of facial pain templates. Apart from facial model-based pain recognition, feature learning-based approaches have also been developed (Mauricio et al., 2019). Zhou et al. (2016) worked on a smooth estimation using a refined Recurrent Convolutional Neural Network (RCNN) framework. The strength of RCNN is that it encodes spatial information (CNN-based advantage) along with temporal information in sequence (RNN-based advantage), which yielded good results regarding both accuracy and running speed on the fully labeled UNBC-McMaster Shoulder Pain Expression Archive database. In addition, Rodriguez et al. (2017) employed a deep learning model with a long-short-term memory (LSTM) to incorporate temporal information into the model. The model outperformed the previous state-of-the-art methods on scores of areas under the curve (AUC), mean standard error (MSE), Pearson correlation coefficient (PCC), and the intraclass correlation coefficient (ICC). Xin et al. (2020) effectively reduced background interference with adaptive weight distribution in facial regions. Bargshady et al. (2020) exploited an Ensemble Deep Learning Model (EDLM) to classify pain and generate a 5-level intensity estimation using the Multimodal database. They also validated the generalizability on the UNBC-McMaster Shoulder Pain dataset.

### 2.2 Voice-based

Despite extensive research on facial cues, few studies have examined how pain can be detected through vocal signals. For humans in the initial developmental stage, sound is more informative to express their feelings (i.e., crying babies) (Tuduce et al., 2018). Research on vocal information has traditionally focused on fundamental problems such as blind source separation (Bell & Sejnowski, 1995; Herault & Jutten, 1986) and speech translation (Jia et al., 2019). A preliminary study found that facial signals can sometimes be independent of voice generation, such as the fundamental frequencies of phonation and intonation (Campanella & Belin, 2007). The separation of voice and facial expressions raises the question of whether we can independently extract pain information from voice or consider voice as a complementary source that benefits pain recognition (see Supplementary Table 3). Literature on computational pain recognition has not yet addressed the voice-based signal. In fact, linguistic and non-linguistic signals can be helpful to the medical service (Besse et al., 2016).

In general, the vocal signal can be divided into two categories (linguistic and non-linguistic), each with distinct characteristics (Noroozi et al., 2018). On the one hand, the linguistic signal is more complex and subjective than the non-linguistic signal, and natural language varies in different contexts (i.e., recreation and work). Currently, social platforms are an integral part of our daily lives, providing an abundance of text data. The abundant learning materials entail labor-saving methods such as semi-supervised learning. In contrast to sentiment analysis, text analysis is less prevalent in computational medical diagnostics. In the long run, such NLP-based technological developments should benefit initial and primary online medical consultations.

On the other hand, information delivery depends on how we use non-linguistic signs—such as pitch, intensity, and tone. Non-linguistic transformations (i.e., Mel-frequency cepstral coefficients, MFCCs; filter bank energies, FBEs; linear spectral pairs, LSP) are critical to extract useful indicators for pain recognition (Noroozi et al., 2017). For example, people can subtly convey opposite information, such as appreciation and sarcasm, by adjusting their tone and stress (Poria et al., 2017). A study found a correlation between bio-signal parameters related to speech prosody and self-reported pain levels (Oshrat et al., 2016). Therefore, "how" we speak (non-linguistic) is sometimes even more important than "what" we say (linguistic). Thus, the extraction of speech features from speech and models for recognizing pain based on speech is crucial to study. A popular audio feature extraction toolkit called OpenSMILE has been proposed to meet the needs of extracting voice features (Eyben, Wöllmer, Graves, et al., 2010; Eyben, Wöllmer, & Schuller, 2010). Additionally, background noise can interfere with voice-based pain recognition. To reduce background noise, researchers need to cleanse the audio information (Hao et al., 2020). Generally, pain recognition through vocal coding is at a very early stage of feature mining compared to sophisticated facial coding systems. A meta-learning process such as stacked ensemble learning may provide insights for speech-based pain recognition (Fig. 2).

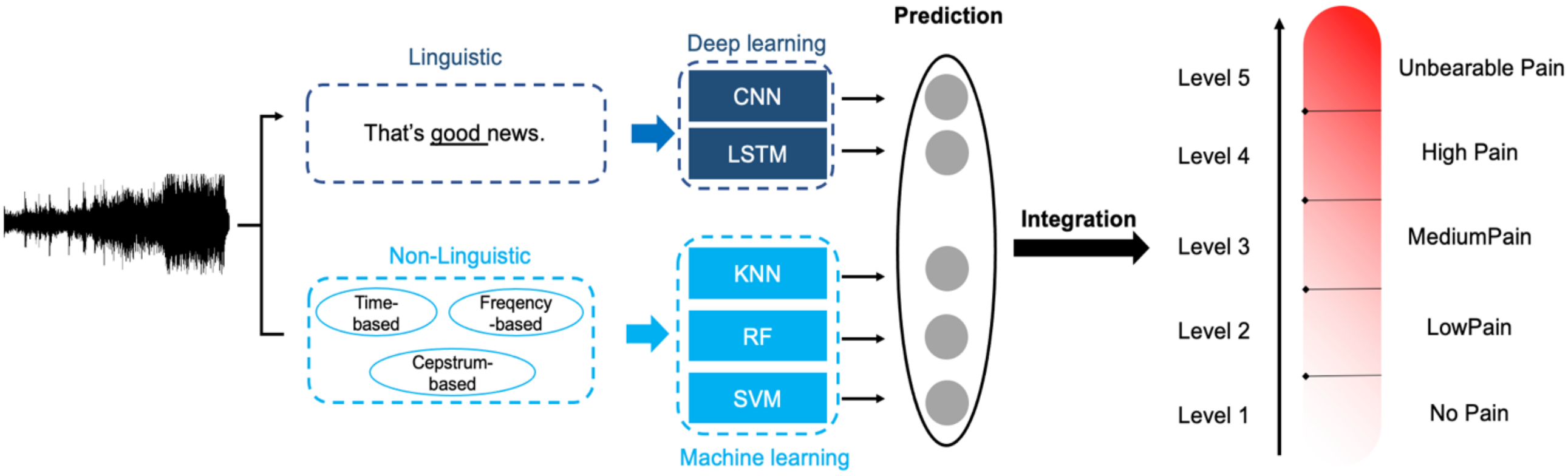

**Fig. 2** A meta-learning process for voice-based pain assessment.

## 2.3 Body-based

Advanced algorithmic models in computer vision relating to human visual systems have driven the development of body-based processing (Parisi et al., 2016). Symbolic movements (such as head nodding) can provide an implicit explanation (Sado et al., 2021). Castellano et al. (2008) suggested that gestures could be the most accurate indicator of pain in a unimodal emotion recognition system, followed by speech and facial expressions. As a result, incorporating body movements into multimodal pain recognition could be beneficial in the long run (Werner et al., 2018). In three databases, studies have examined the head movements and posture patterns of patients in pain- and non-pain states and found remarkable differences, suggesting that body-based pain recognition is feasible (Semwal & Londhe, 2021). In contrast to a large amount of research on face-based pain recognition, few studies have considered body-based pain recognition (see Supplementary Table 4).

The gap in body-based pain recognition could be partly due to the absence of a valid interpretation model for body movement. An analysis of previous research on emotion recognition provides insight into the model of body movement. In particular, two aspects of gestures are examined: propositional (marker-based) and non-propositional (non-selected marker-based) movements. Video sequences capture a variety of body movements, and positional markers are selected from those movements (Stathopoulou & Tsihrintzis, 2011). Noroozi et al. (2018) summarized human body models into two types: part-based and kinematic models. Part-based approaches represent the human body as a flexible configuration of body parts. Kinematic models consist of a series of interconnected joints. However, gestures can be acquired intrinsically (nodding to show approval) or extrinsically (in some cultures waving hands to the side to show rejection instead of greetings). Cultural contexts should be considered in future research to develop an integrated model relevant to the context. For non-positional movement qualities, Castellano et al. (2007) used amplitude, speed, and fluidity of movement to classify eight emotions instead of gestures. Burgoon et al. (2005) investigated context cues and body cues to identify people's emotional states from an abstract perspective based on valence (pleasant/unpleasant) and intensity. However, there is much less research on the choice of either a direct (feature selection based on raw data) or an indirect approach (body models of specific body action units mapping to pain.

Researchers at Carnegie Mellon University (CMU) proposed an OpenPose Body Posture Recognition Project, a real-time multi-person 2D gesture estimation based on deep learning without special hardware to acquire data (Cao et al., 2017). It is an open-source library based on a supervised learning approach and CNN architecture. The project estimates human body movement, facial expression, finger movement, and other body gestures. In contrast to single-target detection, multi-target detection can be achieved with a high level of robustness (Martinez et al., 2019). Despite the difficulty of tracking human pain levels by nonverbal cues such as gestures or postures, various motion capture utilities with depth sensors can record high-quality data, allowing for future progress in this subject (Anbarjafari et al., 2018).

As a whole, development in basic emotion recognition has made significant contributions to affective computing in recent decades. Many achievements have been made in recognizing pain, and this progress is through a generalization of the methodology developed from basic emotion recognition. With the advent of wearable and portable technologies and dry electrode technology, researchers have been able to access additional sources of psychophysiological pain indicators, including electroencephalography (EEG) and Electrodermal Activity (EDA) (i.e., monitoring ICU patients) (Kächele et al., 2015; Walter et al., 2013). One of the most significant aspects of bio-signaling is that physiological parameters may not be correlated with specific emotional states. However, they were related to the underlying emotional representation dimensions (valence, arousal, and dominance) (Schlosberg, 1954). Hence, physiological signals can provide additional information that may be useful in estimating pain intensity (Gruss et al., 2015). A general understanding of the unimodal development of pain recognition presented in this article can assist researchers in developing an optimal multimodal framework or system in the future (Fig.3).

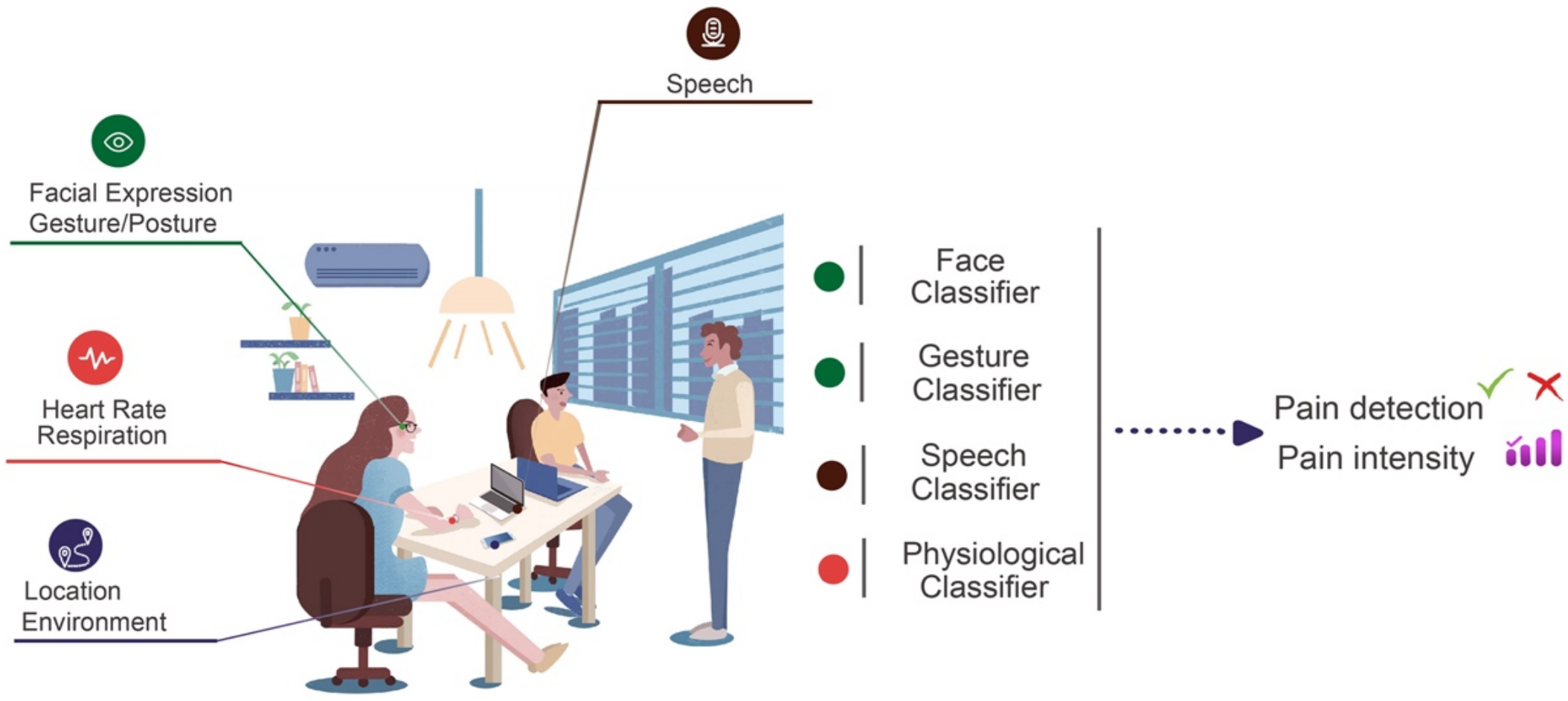

**Fig.3** A sketch of a real-time multimodal pain recognition application

## 3 Cross-disciplinary perspectives on pain recognition

Researchers from computer science have become interested in recognizing human-defined mental processes, including emotions and pain. Pain recognition is not a simple classification problem that entails detecting more than classical pain elements based on physical signals. In addition to serving as a purely spontaneous and bottom-up response to pain stimuli, pain expression also serves a social purpose. Human top-down cognitive systems modulate responses to aversive stimuli. Based on the social function of pain expression, it is possible that physical signals acquired by AI may be confused with social signals related to pain. So, instead of focusing on rapid technological advancement, present research should switch its attention to an understanding of human cognitive, behavioural, and neurological processes.

So far, there has been little attention on whether people can accurately recognize another person's pain and how humans can recognize pain states based on multimodal information, in particular. Studies of human cognition may enrich perspectives for designing new research directions and technology with Human-In-The-Loop (HITL). Under the framework of HITL, this section highlights technical developments based on human cognitive systems, such as multimodal information integration and memory systems, which we regard as a form of indirect learning from humans (Fig. 4 bottom). Furthermore, we propose that computational pain recognition can be improved by learning directly from human studies by focusing on how individuals perceive and respond to the pain of others (Fig. 4 top).

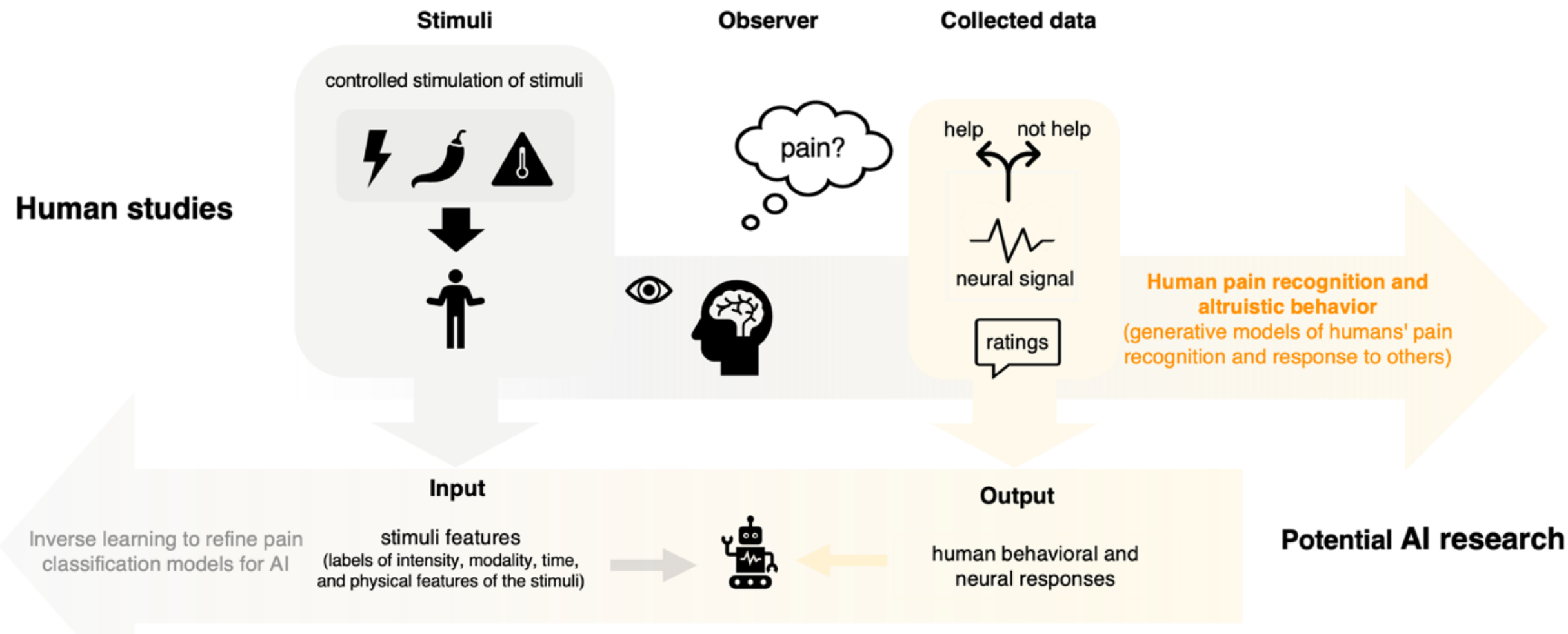

**Fig. 4** Pain recognition under the Human-In-The-Loop framework.

### 3.1 Indirect learning from human neural mechanisms

***Integration and processing of multimodal signals***

It is believed that mental processes are derived from interconnected subsystems in the human brain, and each subsystem contributes to several mental processes (Shallice, 1988). The issue of how algorithms replicate computation in the brain by using reverse engineering is of particular importance, such as the visual system (Hubel & Wiesel, 1959), attention system, and intertwined subsystems (Koch & Ullman, 1987). The brain is able to exchange and match cross-modal information (face and voice) in order to identify others (Blank et al., 2011; Blank et al., 2015). Humans are able to integrate multimodal information by selectively focusing on the relevant inputs that are related to their goals or tasks (Fu et al., 2020). Beyond unimodality, multimodal feature integration methods are rapidly developing (Gruss et al., 2019; Khan et al., 2013; Thiam & Schwenker, 2017), particularly on facial and vocal expressions. Kächele et al. (2015) showed that multimodal information is more advantageous than unimodal in identifying pain from facial expressions, head posture, and physiological signals in videos (see Supplementary Table 5). Unlike unimodal systems, multimodal data fusion has substantially improved recognition accuracy (Ranganathan et al., 2016). Nevertheless, it remains unclear at which stage cross-modal information intersects. The integration of audiovisual speech with cross-modal processing may be in the early phases of nonprimary auditory brain activity (Besle et al., 2008). Also, researchers suggested a hierarchy in which various brain regions encode information independently and interact in the transient cortices (Campanella & Belin, 2007; Davies-Thompson et al., 2019; Molnár et al., 2020). Similarly, the optimal time point for an AI to integrate multimodal information (data fusion) remains controversial.

### *Memory and learning*

Prior knowledge improves our understanding of a situation, and prior experience accelerates our ability to adjust to new situations (Williams & Lombrozo, 2013). Prior experience, such as personal medical history, is a precious source of clinical information. Patients with different types of pain, i.e., acute pain and chronic pain, show a distinct way of processing and expressing pain. For example, researchers have found a bias in information processing when experiencing acute pain compared to chronic pain (Moseley et al., 2005). In this regard, including information about a person in an AI could be beneficial for recognizing pain in a medical context. At the same time, ethical and security standards have to be considered.

Human top-down cognitive systems can modulate pain perception, but significant inter-individual differences exist (Peng et al., 2020). One of the difficulties in pain recognition is the varying expressions towards the same stimuli; that is, humans vary in their expressions in the same situation (Grodal, 2007). Humans have individual differences in expressing pain, so the pattern of responses in prior experiences with similar stimuli is a vital source of learning for pain analysis on specific targets. A personalization of pain intensity estimation systems (Kächele et al., 2017) with multimodal analysis can utilize memory and high-dimensional feature space to represent past information about pain.

Humans rely on the hippocampus as a central hub for memory storage. The hippocampus is activated when an episodic experience replays during resting and sleeping, which has been considered a process that integrates short- and long-term memory (Singer & Frank, 2009). Specifically, researchers proposed an example-based manner by which memory "replays" itself offline to learn the successes or failure cases that occurred in the past (Hassabis et al., 2017). The replay process emphasizes the recirculation of learning. For real-time pain monitoring, a great deal of information from different devices converges in the back-end database. Long-short-term memory (LSTM), inspired by the working memory framework, holds information to a fixed operational state until an acceptable output is required (Hochreiter & Schmidhuber, 1997). LSTMs can contribute to instant detection by allowing the most valuable signals to be processed first and other signals to be processed later.

However, an advanced application would require translating an individual's past experience, pain threshold, or pain display characteristics to a person-specific feature space before training. Over time, data can be trained offline to improve algorithms in specific environments (i.e., hospitals, homes, and public places), similar to human episodic memory, replaying during sleep to realize a conceptual life-long learning system (Sodhani et al., 2022).

### 3.2 Direct learning from human studies

***Modulation of pain expression through social contexts***

From recognition to cognition, individual cognitive patterns (i.e., pain sensitivity) are essential. However, the pattern is context-dependent, meaning pain expressions vary in different contexts. Social contexts can alter an individual's pain expression. For example, our discomfort toward aversive stimuli is lower when we are in the company of others than when we are alone, referring to the social buffering effect (Langford et al., 2006; Qi et al., 2020). Meanwhile, high social threats, such as high electrocutaneous stimuli administered by others, increase pain intensity and unpleasant feelings (Karos et al., 2020). Hammal et al. (2008) suggested that context provides substantial information for pain recognition. These explicit expressions are influenced by our environment. For example, whether people are surrounded by whom they trust (workplace, hospital, or home) (Bublatzky et al., 2020). Therefore, target separation and context are fundamental issues to address for improvement in real-life applications of pain estimation.

***Pain recognition and response to others' pain in human studies***

Other than physical features of pain expressions, what other sources of learning can be integrated into an AI? In most cases, it is beneficial to get inspired and guided by human studies. What is the process by which humans recognize pain in others, and how do they react based on their understanding of the pain state of others? Researchers used well-calibrated experimental pain models (i.e., heat, cold pressure, or video) to generate stimuli that show the pain of others for observers to respond to (Bastian et al., 2014; FeldmanHall et al., 2015). The observers rate the pain level of people in each stimulus, and researchers record their neural signals (EEG, fMRI) of processing these stimuli as well as choices observers make, such as prosocial choices (Jackson et al., 2005; Peng et al., 2021).

Two insights can be gained from some studies concerning human reactions to others in pain. For one thing, stimuli used for eliciting human responses to others' pain have inherent features (standardized labels of others' pain state, and physical features of the stimuli), which are the potential input for any algorithms. In addition, the experiments collected human reactions in several dimensions (behavioral ratings or decisions to help people in pain, which is a feasible output/end for an AI to learn). This line of study can shed light on the "black box" systems of the human cognitive system between the input of victims in pain and the output of observers' reactions. Ultimately, the goal is to explore pain recognition from a human perspective and to allow current computational pain recognition to benefit from human cognitive systems.

### 3.3 Prospects in pain recognition

For the above sections, we first reviewed unimodal- and multimodal- pain recognition systems. We introduced some established databases, followed by an overview of the progress in face-, voice- and gesture-based pain recognition. Regarding robustness, consistency, and implementation prerequisites, all single modalities are restricted. First, there is a shortage of sufficient data, including targets with different pain levels and real-life big clinical data (i.e., spontaneous painful expressions of different modalities). Also, realistic scenarios are often ignored. Second, multimodal pain models ought to be prepared on substantial information from various contexts to construct generalizable models. Third, a meta-analysis of diverse methods regarding specific modalities provides insights into pain recognition. For example, what information can be weighted more than the other to improve performance? How can recognition accuracy be enhanced by combining the outputs of different independent models? Specifically, ensemble learning, which combines several weak classifiers into one integrated classifier, is a potential solution for building better models of pain analysis. Moreover, it is also a way to address how to deal with multisensory information processing, and how intelligent recognition can be done in the real world with more available signals in the future (Fig. 3).

## 4 Beyond pain recognition

In retrospect, pain is often associated with empathy and noble virtues, such as altruism, from a psychological perspective (Cameron et al., 2019; Fehr & Fischbacher, 2003; Staub & Vollhardt, 2008). Empathic people can recognize and comprehend others' emotions by experiencing and sharing the emotional states of others (Singer, 2006). However, the substantial relationship between pain and empathy has not been explored in AI research. An AI agent with human-like empathic responses is considered more caring, likeable, and trustworthy (Rodrigues et al., 2015). In the current state-of-the-art, one question is, "can an AI express empathy at all?". There is still a long way to go before we get closer to this question. However, could the recognition of pain, which has relatively clear and definite features for an AI agent to learn, lay the foundation for artificial pain empathy? There are two steps for an AI to implement pain empathy: first, the identification of pain, and second, the expression of empathy (Khatibi & Mazidi, 2019). The above sections have reviewed the pain recognition progress. In this section, we describe a relatively novel field associated with an affective AI — artificial pain empathy.

## 4.1 Artificial pain empathy

***Mimicry***

Pain empathy is the capacity associated with feeling and evaluating others' pain states and understanding others, often prompting prosocial actions (De Waal, 2008; Fitzgibbon et al., 2010; Wang et al., 2019). Studies from animals and humans suggested that mimicry of body movement is the underlying mechanism for empathy which derives from a mirror system in the human brain (Rizzolatti & Craighero, 2004). In social interactions, emotional mimicry is crucial since it reflects a desire to connect with another person (Hess & Fischer, 2013). Thus, the initial step to addressing pain empathy is to recognize and imitate human facial expressions or gestures in real-time. To some extent, mimicry generates similarity and, in turn, facilitates people's empathy toward a human-designed machine (Breazeal, 2003). Miura et al. (2008) found that human-like body movements make it easier for people to empathize and the embodiment of an interactive partner influences human mimicry behaviors. A physical presence and human-like artificial entity tend to generate more mimicry than virtual non-human counterparts (Perugia et al., 2020). In addition, emotional mimicry by robots may show its empathic "trait", improving the human-robot interaction experience (Leite et al., 2012).

***Modulation by top-down cognitive and presence of social partners***

What functions can AI agents or robots have when they cannot feel suffering like humans? First**,** some insights can be gained from studies on people born with the congenital absence of pain. Danziger et al. (2009) found that pain-related brain regions of congenitally pain-free patients are activated when seeing others experiencing pain, indicating a shared synchrony pain aversion with others. Krishnan et al. (2016) found that the experience of vicarious pain (observing others in pain) is neurologically separate from experiencing actual pain on our own, suggesting that empathy is more cognitive than sensational. Thus, empathy not only has the intrinsic characteristics of sensibility but also comprises the process of top-down cognitive regulation. Heyes (2018) proposed a dual-system model of empathy. The model includes both early views in which empathy largely depends on a bottom-up process, a spontaneous response activated by stimuli (system I). Meanwhile, it also covers the control mechanism of empathy that belongs to a top-down process, indicating that high-level cognitive systems contribute to the regulation of empathy (system II). The two-system model of empathy lays the foundation for the investigations on artificial empathy from a cognitive perspective.

Second, the social buffering effect indicates that the presence of social partners effectively modulates human reactions toward aversive stimuli (Qi et al., 2020). A recent study on rodents found a neural circuit of the buffering effect. Their results showed that brief social interaction with a peer mouse experiencing pain or morphine analgesia resulted in the transfer of these experiences to its social partner (Smith et al., 2021). Hence, a theoretical hypothesis is suggested that painless robots can serve as analgesic companions, thereby reducing people's pain perception. In fact, mental support is more human-centered and user-friendly than realizing high-level pain recognition. Some effort has been put into this practice. For example, the Institute for Creative Technologies at the University of Southern California created an empathic AI system that functions as a virtual counselor, and mainly serves veterans with post-traumatic stress disorder (Gaggioli, 2017).

Furthermore, when humans suffer together, they prefer to bond (Peng et al., 2021). Thus, the pain-sharing demonstration of artificial agents increases the possibility of building trust with people. In addition, practices in human studies showed that neuromodulation, an array of non-invasive, minimally invasive, and surgical electrical therapies, is conducive to pain relief (Knotkova et al., 2021). People can self-adjust their feelings by training them with positive feedback when making target responses, and the technique is called neurofeedback. Researchers have found evidence that the sensory-discriminative aspect of pain is associated with EEG signals that are deliberatively trained. However, due to differences in training efficacy, future research is needed to examine individualized neurofeedback for pain regulation (Peng et al., 2020).

### 4.2 Empathy training for humans

Another potential application for AI research to contribute to HILP systems is to use a generative model. The generative models that generate human images and speech have come close to equivalents in the real-world (Oord et al., 2016). On the one hand, synthetic data can be used to train the recognition models. On the other hand, the data can be applied to the training process for medical caregivers to improve people's estimation of pain intensity. A recent study indicated the feasibility of such training pain. People underwent a 3.5 to 5 hours online training program called the index of facial pain expression (IFPE). After the short training, observers improved their identification of others' psychiatric pain expressions (Rash et al., 2019). Hence, an important task for future artificial agents is to help people with emotional deficiency to establish the ability to perceive, interpret, express, and regulate emotions (Javed & Park, 2019).

To sum up, artificial assistants for pain modulation are becoming available to assist humans. The presence of them and simple pain-sharing displays can have a significant effect in reducing pain perception. More sophisticated neuromodulation is also feasible in the future application of artificial medical assistants. To jointly contribute to the field, it is required to integrate psychology, ethics, and cognitive neuroscience.

### 4.3 Ethical considerations

Concerning the overall topic, ethical issues should be considered with attention and responsibility for future work on affective assistants. Ideally, computational pain recognition can be applied to assist people in pain. Artificial empathy refers to the capacity of computer systems to recognize and respond to people's behaviors, expressions, and emotions (Asada, 2015a, 2015b; Asada et al., 2012). With each interaction people have with an AI system, adaptive artificial empathy can be expected (Harris &

Sharlin, 2010). In this paper, we reviewed some existing databases, which may be helpful for developing algorithms that can generate empathic expressions. Furthermore, an AI agent needs to recognize and generate an empathic response to others' pain experiences. However, research has not yet fully considered an affective AI with proper responses to others' pain experiences. By learning from human-human interaction, an AI system could also learn human behaviors by interacting with real people in an "empathic" way (Paiva et al., 2017).

Although there is a promising future for artificial systems with empathy-like responses, the concept is more than modeling humans (James et al., 2018; Putta et al., 2022; Srinivasan & San Miguel González, 2022). There are important issues to be considered: What artificial systems could be advantageous and beneficial for human life? What boundaries do such service functions need to avoid crossing? Generally, the precision, flexibility, and convenience of advanced services require personal information from users. Therefore, the privacy and security of personal data need to be considered highly. Rules for accountability require legal support and government supervision. More significantly, if artificial systems are launched, both the creator and the user must be fully informed about the information exchanged, possible health concerns and the limitations of interpreting affect. Trustworthiness and understanding of AI systems' limitations, like algorithmic bias by biased data, are important topics to address, particularly when people interact with artificial empathy services.

### 5 Conclusion

This paper highlights the potential of multimodal signals to improve pain recognition in computational models and assistive systems. Regular pain monitoring by assistive AI systems may allow patients to obtain timely treatment, and an AI assistant may be helpful for human-computer interaction. However, challenges include noise estimation in unimodal sources, multiple feature extraction, comparison of fusion methods, and establishing balanced and compatible datasets to address the need for real-time applications. Moreover, an AI assistant should show socially acceptable verbal and non-verbal signals to achieve positive interactions. Although AI can demonstrate empathic responses, there are inherent limitations and issues regarding the semantic understanding of intention and trust when processing affective information. Integrating and exploring affective AI jointly by psychologists and computer scientists will help better interpret how humans understand each other based on sensory and emotional states. If pain recognition and human-like empathic expressions are developed in an ethical manner, an artificial agent could also be considered a helpful assistant trained to express empathy and safely interact with humans.

# Supplementary Information

## Can AI detect pain and express pain empathy? A review from emotion recognition and a human-centered AI perspective

**Running head:** Pain Recognition and Empathic Expression in AI

**Table 1** Pain databases

| Database | Type | Modality | Subjects | Reference | Description |
| --- | --- | --- | --- | --- | --- |
| UNBC-McMaster Shoulder Pain Expression Archive (Lucey et al., 2011) | spontaneous | video | 129 adults shoulder pain patients | Lucey et al. (2011) | All videos record the faces of participants with shoulder pain. Participants are instructed to perform a sequence of tests that require motions of limbs in two different scenarios. |
| STOIC database (Roy et al., 2007) | acted | video | 10 actors (age 20-45) | Roy et al. (2007) | All videos include basic emotions, painful and neutral expressions. |
| EmoPain (Aung et al., 2015) | spontaneous | video, audio, sEMG | 50 subjects (22 chronic low back pain patients and 28 healthy subjects with no history of chronic low back pain) | Aung et al. (2015) | A multimodal dataset that is completely marked. |
| Binghamton–Pittsburgh 4D Spontaneous Facial Expression Database (BP4D) (Zhang et al., 2014) | elicited | videos | 41 healthy adults (age 18-29) | Zhang et al. (2014) | Social or non-social stimuli are used to elicit pain. All facial expressions are recorded during natural social contact. |
| The "Multimodal Intensity Pain" database (Mintpain) (Haque et al., 2018) | elicited | video | 20 healthy adults (age 22-42) | Haque et al. (2018) | Painful expressions of participants are elicited by electrical pain stimulation. Each video frame is labeled with five pain levels. |
| The Biovid heat pain database (BioVid) (Walter et al., 2013) | elicited | video, SCL, ECG, EMG, EEG | 90 healthy adults (age 20-65) | Walter et al. (2013) | The pain stimuli that participants received is customized. Total five levels of pain intensity are annotated. |
| The Infant COPE Database (Brahnam et al., 2006) | spontaneous elicited | image | 26 neonates (age 18-36 hours) | Brahnam et al. (2006) | Twenty-six neonates experience the pain of heel lancing as well as three non-pain stressors. |
| YouTube Dataset (Harrison et al., 2014) | elicited | video | 142 infants (age 0-12 months) | Harrison et al. (2014) | Face, body, and sounds are recorded. |
| The SenseEmotion Database (Velana et al., 2017) | elicited | video, audio, SCL, ECG, EMG, RSP | 45 healthy subjects | Velana et al. (2017) | Pain is induced by heat stimulation. Five classes from no pain to pain. |
| X-ITE Pain Database (Gruss et al., 2019) | elicited | video, ECG, SCL, EMG | 134 healthy adults (age 18-50) | Gruss et al. (2019) | Specific pain thresholds (from low to intolerable pain) in the record are determined based on the previous calibration. |
| Duesseldorf Acute Pain (DAP) Corpus (Ren et al., 2018) | elicited | audio | 80 subjects (age 18-70) | Ren et al. (2018) | Pain is induced using a cold pressor test while participants perform different reading and free-form speech tasks. |
| The iCOPEvid dataset (Brahnam et al., 2020) | elicited | video | 49 neonates | Brahnam et al. (2020) | A series of neonatal facial expressions videos are included. |

Skin Conductance Level (SCL), Electrocardiogram (ECG), Electromyogram (EMG), surface Electromyographic study (sEMG), Respiration (RSP), and Electroencephalogram (EEG).

**Table 2** Pain recognition based on facial expression

| References | Database | Feature descriptor | Learning method | Goal | Classifier | P |
|---|---|---|---|---|---|---|
| Lucey et al. (2009) | UNBC | AAM landmarks, AUs (Lucey et al., 2009) | Supervised | pain/no-pain | SVM | 7 |
| Lucey et al. (2011) | UNBC | SPTS, CAPP (Lucey et al., 2012) | Supervised | pain/no-pain | SVM | |
| Florea et al. (2014) | UNBC | AAM landmarks, HoT (Florea et al., 2014) | Supervised | pain/no-pain | SVR, transfer learning | |
| Kaltwang et al. (2012) | UNBC | AAM landmarks, DCT, LBP (Kaltwang et al., 2012) | Supervised | pain intensity | RVR | |
| Chen et al. (2012) | UNBC | LBP (Chen et al., 2012) | semi-supervised | pain/no-pain | Transfer learning | |
| Khan et al. (2013) | X-ITE | PLBP, PHOG (Khan et al., 2013) | Supervised | pain/no-pain | SVM | |
| Pedersen (2015) | UNBC | AAM landmarks, autoencoder (Pedersen, 2015) | Supervised | pain/no-pain | SVM | |
| Zhou et al. (2016) | UNBC | AAM-warped frame vector sequences (Zhou et al., 2016) | Supervised | pain intensity | RCNN regression | |
| Zhao et al. (2016) | UNBC | self-learned features (Zhao et al., 2016) | supervised, semi-supervised, unsupervised | pain intensity | Ordinal Support Vector Regression | |
| Rodriguez et al. (2017) | UNBC | VGG_Faces features (Rodriguez et al., 2017) | Supervised | pain intensity | CNN-LSTM | |
| Xin et al. (2020) | UNBC | self-learned features (Xin et al., 2020) | Supervised | pain intensity | CNN | |
| Bargshady et al. (2020) | MIntPAIN<br>UNBC | VGG_Faces features (Bargshady et al., 2020) | Supervised | pain intensity | Ensemble Deep Learning Model | |

Similarity Normalized Shape or Points (SPTS); Normalized Appearance (SAPP); Canonical Normalized Appearance (CAPP); Histogram of Topographical features (HoT); Discrete Cosine Transform (DCT); Local Binary Patterns (LBP); Second-order standardized moment average pooling (2Standmap); Pyramid Local Binary Patterns (PLBP); Pyramid Histogram of Orientation Gradients (PHOG)

**Table 3 Pain recognition based on voice**

| References | Database | Classifier | Accuracy | CCI |
|---|---|---|---|---|
| Tsai et al. (2017) | Collected themselves (Tsai et al., 2017) | DBFs+LSTM | 72.3% | — |
| Oshrat et al. (2016) | Collected themselves (Oshrat et al., 2016) | SVM | | 0.74 |
| Hossain (2016) | Collected themselves (Hossain, 2016) | GMM | 92.4% | — |
| Tuduce (2018) | Baby Cry Database (Tuduce et al., 2018) | 30 classifiers in WEKA | 70% | — |

Gaussian Mixture Model (GMM); Waikato Environment for Knowledge Analysis (WEKA); Deep Bottleneck Features (DBFs).

## Table 4 Pain recognition based on body

| References | Database | Classifier | Accuracy |
|---|---|---|---|
| Olugbade et al. (2014) | Collected themselves (Olugbade et al., 2014) | SVM | 63% |
| Saha et al. (2015) | Collected themselves (Saha et al., 2015) | K-NN | 90.63% |
| Olugbade et al. (2015) | Emo-Pain (Olugbade et al., 2015) | two-level SVM | 94% |
| Saha et al. (2016) | Collected themselves (Saha et al., 2016) | An interval type-2 fuzzy logic-based classifier | 92.14% |

**Table 5** State of the art of multimodal pain recognition

| References | Database | Modality | Fusion Type | Classifier | Accuracy |
|---|---|---|---|---|---|
| Thiam and Schwenker (2017) | SenseEmotion | F+A+Bio+P (Thiam & Schwenker, 2017) | early fusion/late fusion | Random Forest | 61/85% |
| Werner et al. (2019) | X-ITE | F+A+Bio (Werner et al., 2019) | early fusion/late fusion | Random Forest | 74.4%/76.8% |
| Kächele et al. (2015b) | BioVid Heat Pain | F+A+Bio (Kächele et al., 2015) | early fusion/late fusion | Random Forest | 78.9%/77.4% |
| Haque et al. (2018) | MIntPAIN | F(RGBDT) (Haque et al., 2018) | early fusion/late fusion | CNN + LSTM | 36.55%/25.4% |
| Kessler et al. (2017) | SenseEmotion | F+Bio (Kessler et al., 2017) | hierarchical fusion | Random Forest | 71.85% |
| Salekin et al. (2021) | Recorded in hospital | F+A+G (Salekin et al., 2021) | Late fusion | VGG-16+CNN+LSTM | 79% |
| Rivas xet al. (2020) | Recorded in hospital | F+FP+HM (Rivas et al., 2021) | Late fusion | SNBC+CCC | 90.5% |
| Badura et al. (2021) | Collected themselves | Bio (Badura et al., 2021) | Late fusion | AdaBoost | 84% |

Face (F); audio (A); Bio-physiology (Bio); RGB Depth and Thermal (RGBDT); Geature (G); Finger pressure (FP); Hand movements (HM); Semi-Naive Bayesian Classifier (SNBC), Circular Classifier Chain (CCC).